\documentclass[11pt]{article}

\usepackage{microtype,palatino}
\usepackage{graphicx,xcolor}
\usepackage{booktabs} 
\usepackage{times}
\usepackage{epsfig}
\usepackage{amsmath, amssymb, comment}
\usepackage{algpseudocode,algorithm,algorithmicx}
\usepackage[labelfont=bf,list=true,font=footnotesize]{subcaption}
\usepackage[margin=0pt,font=footnotesize,labelfont=bf]{caption}
\usepackage{float, enumerate,enumitem}
\usepackage{flushend}
\usepackage{soul}
\usepackage{multirow}

\DeclareMathOperator*{\argmin}{arg\,min}

\newcommand{\Redit}[1]{ {#1}}
\newcommand{\Rdelete}[1]{}

\usepackage{hyperref}
\hypersetup{
	colorlinks = true,
	urlcolor = blue,
	filecolor = blue,
	linkcolor=black, 
	citecolor=black, 
}

\date{}

\begin{document}

\begin{center}

{\bf{\LARGE{
Generative Models for Low-Rank Video Representation and Reconstruction
}}}

\vspace*{.2in}

{\large{
\begin{tabular}{cccc}
Rakib Hyder and M. Salman Asif\\
\end{tabular}
}}

\vspace*{.05in}

\begin{tabular}{c}
Department of Electrical and Computer Engineering, University of California, Riverside\\
\end{tabular}

\vspace*{.1in}


\end{center}

\begin{abstract}
Finding compact representation of videos is an essential component in almost every problem related to video processing or understanding. In this paper, we propose a generative model to learn compact latent codes that can efficiently represent and reconstruct a video sequence from its missing or under-sampled measurements. We use a generative network that is trained to map a compact code into an image. We first demonstrate that if a video sequence belongs to the range of the pretrained generative network, then we can recover it by estimating the underlying compact latent codes. Then we demonstrate that even if the video sequence does not belong to the range of a pretrained network, we can still recover the true video sequence by jointly updating the latent codes and the weights of the generative network. 
To avoid overfitting in our model, we regularize the recovery problem by imposing low-rank and similarity constraints on the latent codes of the neighboring frames in the video sequence. We use our methods to recover a variety of videos from compressive measurements at different compression rates. We also demonstrate that we can generate missing frames in a video sequence by interpolating the latent codes of the observed frames in the low-dimensional space. 
\end{abstract}

\section{Introduction}

\begin{figure}[b]
\centering

\includegraphics[width=0.5\textwidth]{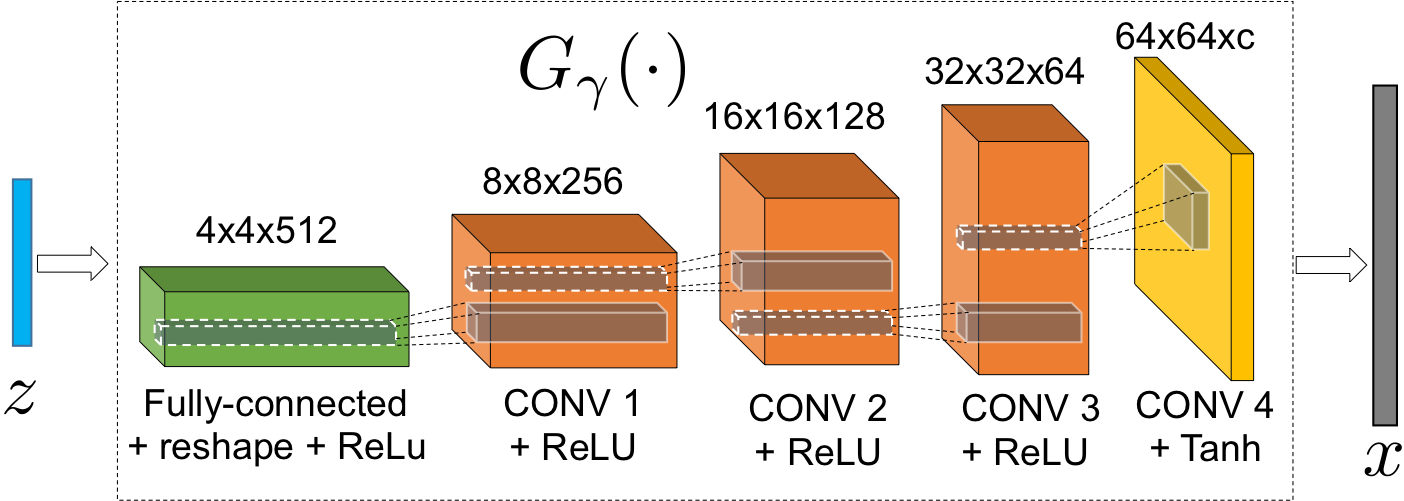}
\caption{Generative model: $x = G_\gamma(z)$ maps a vector $z\in\mathbb{R}^k$ into an image $x\in \mathbb{R}^n$. The figure shows a DCGAN architecture that we used in our experiments with one fully connected and four convolutional layers. }
\label{fig:gen}
\end{figure}

Deep generative networks, such as autoencoders, generative adversarial networks (GANs), and variational autoencoders (VAEs), are now commonly used in almost every machine learning and computer vision task \cite{goodfellow2014generative,ronneberger2015u,karras2017progressive,vondrick2016generating}. One key idea in these generative networks is that they can learn to transform a low-dimensional feature vector (or latent code) into realistic images and videos. The \textit{range} of the generated images is expected to be close to the true underlying distribution of training images. Once these networks are properly trained (which remains a nontrivial task), they can generate remarkable images in the trained categories of natural scenes. 

In this paper, we propose to use a deep generative model for compact representation and reconstruction of videos from a small number of linear measurements. We assume that a generative network trained on some class of images is available, which we represent as 
\begin{equation}\label{eq:genModel}
    x = G_\gamma(z) \equiv g_{\gamma_L}\circ g_{\gamma_{L-1}} \circ \cdots \circ g_{\gamma_1}(z).
\end{equation}
$G_\gamma(z)$ denotes the overall function for the deep network with $L$ layers that maps a low-dimensional (latent) code $z \in \mathbb{R}^k$ into an image $x \in \mathbb{R}^n$ and $\gamma = \{\gamma_1,\ldots, \gamma_L\}$ represents all the weight parameters of the deep network. $G_\gamma(\cdot)$ as given in \eqref{eq:genModel} can be viewed as a cascade of $L$ functions $g_{\gamma_l}$ for $l=1,\ldots,L$, each of which represents a mapping between input and output of respective layer. An illustration of such a generator with $L=5$ is shown in Figure~\ref{fig:gen}. 
Suppose we are given a sequence of measurements for $t=1,\ldots,T$ as 
\begin{equation}\label{eq:measModel}
    y_t = A_t x_t + e_t,
\end{equation}  
where $x_t$ denotes the $t^{th}$ frame in the unknown video sequence, $y_t$ denotes its observed measurements, $A_t$ denotes the respective measurement operator, and $e_t$ denotes noise or error in the measurements. Our goal is to recover the video sequence ($x_t$) from the available measurements ($y_t$). The recovery problem becomes especially challenging as the number of measurements (in $y_t$)  becomes very small compared to the number of unknowns (in $x_t$). To ensure quality reconstruction in such settings, we need a compact (low-dimensional) representation of the unknown signal. Thus, we use the given generative model to represent the video sequence as 
$x_t = G_\gamma(z_t)$ and the observed measurements as $y_t = A_t G_\gamma(z_t)$. 

We first demonstrate that if a video sequence $(x_t)$ belongs to the range of the network $G_\gamma(z_t)$, then we can reconstruct it by optimizing directly over the latent code $z_t$. Then we demonstrate that even if a video sequence lies outside the range of the given network $G_\gamma(z_t)$, we can still reconstruct it by jointly optimizing over network weights $\gamma$ and the latent codes $z_t$. To exploit similarities among the frames in a video sequence, we also include low-rank and similarity constraints on the latent codes. We note that the pretrained network we used in our experiments is highly overparameterized; therefore, low-rank and similarity constraints help in regularizing the network and finding good solution presumably near the initial weights. 

\subsection{Motivation and Related Work}
Video signals have natural redundancies along spatial and temporal dimensions that can be exploited to learn their \textit{compact} representations. Such compact representations can then be used for compression, denoising, restoration, and other processing/transmission tasks. Historically, video representation schemes have relied on hand-crafted blocks that include motion estimation/compensation and sparsifying transforms such as discrete cosine transform (DCT) and wavelets \cite{wallace1992jpeg,christopoulos2000jpeg2000,sullivan2004h,puri1998mpeg}. Recent progress in data-driven representation methods offers new opportunities to develop improved schemes for compact representation of videos \cite{pan2016hierarchical,santurkar2018generative,lin2016deep}.  

Compressive sensing refers to a broad class of problems in which we aim to recover a signal from a small number of measurements \cite{candes2011compressed, donoho2006compressed, candes2005decoding}. The canonical compressive sensing problem in \eqref{eq:measModel} is inherently underdetermined, and we need to use some prior knowledge about the signal structure. Classical signal priors exploit sparse and low-rank structures in images and videos for their reconstruction \cite{duarte2008single,baraniuk2007compressive, yang2013adaptive,shi2012video,zhao2017video}.

Deep generative models offer a new framework for compact representation of images and videos. A generative model can be viewed as a function that maps a given input (or latent) code into an image. For compact representation of images, we seek a generative model that can generate a variety of images with high fidelity using a very low-dimensional latent code. Recently, a number of generative models have been proposed to learn latent representation of an image with respect to a generator \cite{lipton2017precise, zhu2016generative,creswell2018inverting}. %
The learning process usually involves gradient decent to estimate the best representation of the latent code, where the gradients with respect to the latent code representation are backpropagated to the pixel space \cite{Bojanowski2018OptimizingTL}. 

In recent year, generative networks have been extensively used for learning good representations for images and videos. Generative adversarial networks (GANs) and variational autoencoders (VAEs) \cite{goodfellow2014generative,kingma2013auto,hinton2006fast,bengio2013representation} learn a function that maps vectors drawn from a certain distribution in a low-dimensional space into images in a high-dimensional space. An attractive feature of VAEs \cite{kingma2013auto} and GANs \cite{goodfellow2014generative} is their ability to transform feature vectors to generate a variety of images from a different set of desired distributions. Our technical approach bears some similarities with recent work on image generation and manipulation via conditional GANs and VAEs \cite{chen2016infogan, gulrajani2017improved, shrivastava2017learning}. For example, we can create new images with same content but different articulations by changing the input latent codes \cite{chen2016infogan, radford2015unsupervised}.
In \cite{Bojanowski2018OptimizingTL}, the authors presented a framework for jointly optimizing latent code and network parameters while training a standalone generator network. Furthermore, linear arithmetic operations in the latent space of generators can generate to meaningful image transformations. In our paper, we will apply similar principles to generate different frames in a video sequence while jointly optimizing latent codes and generator parameters but ensuring that latent codes belong to a small subspace (even a line as we show in Figure~\ref{fig:2dEmbedding}).

In this paper, we use a generative model as a prior for video signal representation and reconstruction. Our generative model and optimization is inspired by recent work on using generative models for compressive sensing in  \cite{bora2017compressed,van2018compressed,heckel2018deep,shah2018solving,Ulyanov2017DeepIP}. 
Recently, \cite{bora2017compressed} showed that a trained deep generative network can be used as a prior for image reconstruction from compressive measurements; the reconstruction problem involves optimization over the latent code of the generator. In a related work, \cite{Ulyanov2017DeepIP} observed that an untrained convolutional generative model can also be used as a prior for solving inverse problems such as inpainting and denoising because of their tendency to generate natural images; the reconstruction problem involves optimization of generator network weights. Inspired by these observations, a number of methods have been proposed for solving compressive sensing problem by optimizing generator network weights while keeping the latent code fixed at a random value \cite{heckel2018deep, van2018compressed}. As they are allowing generator parameters to change, the generator can reconstruct wide range of images. However, as the latent codes are initialized randomly and stay the same, we cannot find a representative latent codes for images.

\begin{figure}
\centering 
\begin{subfigure}[b]{0.25\linewidth}
\centering
\includegraphics[width=1\textwidth]{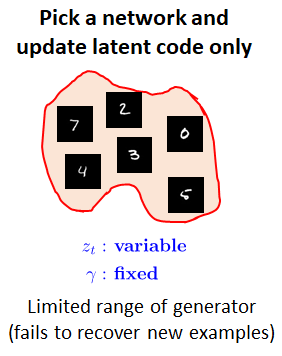}
\caption{}
\end{subfigure}
~~~
\begin{subfigure}[b]{0.28\linewidth}
\centering
\includegraphics[width=1\textwidth]{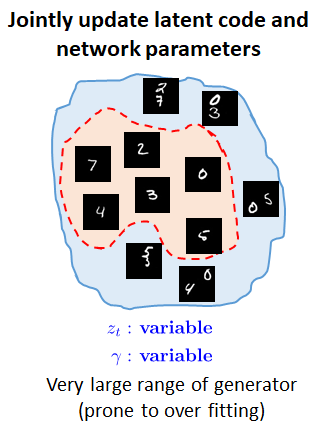}
\caption{}
\end{subfigure}
~~~
\begin{subfigure}[b]{0.27   \linewidth}
\centering
\includegraphics[width=1\textwidth]{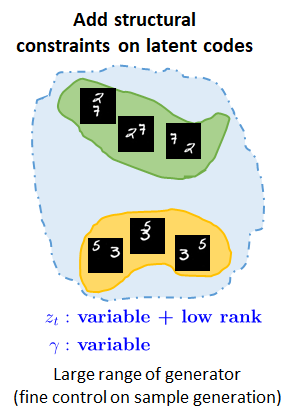}
\caption{}
\end{subfigure}
\caption{An illustration of different generative models discussed in the paper: (a) Optimizing latent codes can only reconstruct images in the range of the generative network. (b) Jointly optimizing latent code and network weights enables recovery of a larger range of images. (c) Low-rank and similarity constraints on latent code  further regularize the problem and potentially explain other structures in data.} \label{fig:intro}
\end{figure}

In our proposed method, we use the generative model in \eqref{eq:genModel} to find compact representation of videos in the form of $z_t$. To reconstruct a video sequence from the compressive measurements in \eqref{eq:measModel}, we either optimize over the latent codes $z_t$ or or optimize over the network weights $\gamma$ and $z_t$ in a joint manner. Since the frames in a video sequence exhibit rich redundancies in their representation. We hypothesize that if the generator function is continuous, then the similarity of the frames would translate into the similarity in their corresponding latent codes. Based on this hypothesis, we impose similarity and low-rank constraints on the latent codes to represent the video sequence with an even more compact representation of the latent codes. An illustration of the differences between the types of representations is shown in Figure~\ref{fig:intro}.

\subsection{Main Contributions}
In this paper, we propose to use a low-rank generative prior for compact representation of a video sequence, which we then use to solve some video compressive sensing problems. The key contributions of this paper are as follows. 

\begin{itemize}[leftmargin=*]
\item We first demonstrate that we can learn a compact representation of a video sequence in the form of low-rank latent codes for a deep generative network similar to the one depicted in Figure~\ref{fig:gen}. 

\item Consecutive frames in a video sequence share lots of similarities. To encode similarities among the reconstructed frames, we introduce low-rank and similarity constraints on the generator latent codes. This enables us to represent a video sequence with a very small number of parameters in the latent codes and reconstruct them from a very small number of measurements. 
 
 \item Latent code optimization can only reconstruct a video sequence that belong to its range. We demonstrate that by jointly optimizing the latent codes with the network weights, we can expand the range of the generator and reconstruct images that the given initial generator fails on. We show that even though the network has a very large number of parameters, but the joint optimization still converges to a good solution with similarity and low-rank constraints on latent codes.

\item We show that, in some cases, the low-rank structure on the latent codes also provides a nice low-dimensional manifold that can be used to generate new frames that are similar to the given sequence. 
\end{itemize}

\section{Technical Approach}

Let us assume that $x_t\in \mathbb{R}^{n}$ for $t=1,\ldots, T$ is a sequence of video frames that we want to reconstruct from the measurements $y_t = A_t x_t + e_t$ as given in \eqref{eq:measModel}. 
The generative model as given in \eqref{eq:genModel} maps a low-dimensional representation vector, $z_t \in \mathbb{R}^k$, to a high-dimensional image as $x_t = G_\gamma(z_t)$. Thus, our goal of video recovery is equivalent to solving the following optimization problem over $z_t$: 
\begin{equation}\label{eq:measModel2} 
    y_t = A_t G_\gamma(z_t) + e_t,
\end{equation}
which can be viewed as a nonlinear system of equations. 

\subsection{Latent Code Optimization}

In latent code optimization, we assume that the function $G_\gamma(\cdot)$ approximates the probability distribution of the set of natural images where our target image belongs. Thus, we can restrict our search for the underlying video sequence, $x_t$, only in the range of the generator. Similar problem has been studied in \cite{bora2017compressed} for image compressive sensing.

Given a pretrained generator, $G_\gamma$, measurement sequence, $y_t$, and the measurement matrices, $A_t$, we can solve the following optimization problem to recover the low-dimensional latent codes: $\hat z_t$ for our target video sequence, $\hat x_t = G_\gamma(\hat z_t)$, as
\begin{equation}
     \hat z_1,\ldots \hat z_T = \argmin_{z_1,\ldots,z_T}\;\sum_{t=1}^T\|y_t-A_t G_\gamma(z_t)\|_2^2. \label{eq:latentOpt}
\end{equation}

Since we can backpropagate gradient w.r.t. the $z_t$ through the generator, we can solve the problem in \eqref{eq:latentOpt} using gradient descent. Although latent code optimization can solve compressive sensing problem with high probability, it cannot solve the problem when the images do not belong to the generator. As there are wide variety of images, it is difficult to represent them with a single or a few generators. In such scenarios, latent code optimization proves to be inadequate.

\subsection{Joint Optimization of Latent Codes and Generator}
\begin{figure}[t]
\centering
    
   \includegraphics[width=0.55\textwidth]{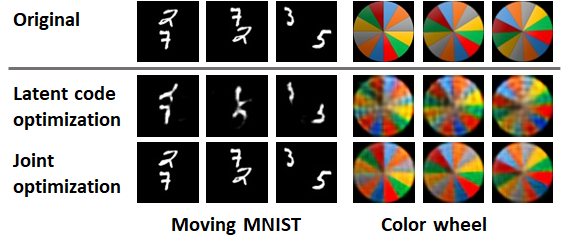}
    \caption{Comparison between optimization over latent code alone and joint optimization over latent code and network weights for representing Moving MNIST and Color wheel test sequence. Latent code optimization fails to generate quality images of sequences that are very different from the training set. Joint optimization can generate both sequences because we have very large degrees of freedom.}

   \label{mov_mnist}
   
\end{figure}

Any generator has a limited range within which it can generate images; the range of a generator presumably depends on the types of images used during training. 
To highlight this limitation, we performed an experiment in which we tried to generate a video sequence that is very different from the examples on which our generator was trained on. This is not a compressive sensing experiment; we are providing original video sequences $x_t$ to the generator and finding the best approximation of the sequence generated by them. The results are shown in Figure~\ref{mov_mnist} using two video sequences: Moving MNIST and Color Wheel. In both cases, network weights are initialized with the weights of a generator that was trained on a different dataset. The pretrained network used for Moving MNIST example was trained on standard MNIST dataset, which does not include any image with two digits. Therefore, the generator trained on MNIST fails on Moving MNIST if we only optimize over the latent code because Moving MNIST dataset consists of images with two digits. The joint optimization of latent code and generator parameters, however, can recover the entire Moving MNIST sequence with high quality. For Color wheel the original generator was trained on CIFAR10 training set which contains diverse category of images. However, as we see in Figure~\ref{mov_mnist}, the generator fails to produce quality images Still it cannot perform well on color wheel representation just by latent code update. Joint optimization improves the reconstruction quality significantly. 

The results presented in Figure~\ref{mov_mnist} should not be surprising for the following reasons: We are providing a video sequence $x_t$ to the generator $G_\gamma(z_t)$ that has $k$ degrees of freedom for each $z_t$; therefore, the range of sequences that can be generated by changing the $z_t$ is quite limited for a fixed $\gamma$. In contrast, if we let $\gamma$ change while we learn the $z_t$, then the network can potentially generate any image in $\mathbb{R}^n$ because we have a very large degrees of freedom. Note that in our generator, the number of parameters in $\gamma$ is significantly larger than the size of $x_t$ or $z_t$. 

The surprising thing, however, is that we can also recover quality images by jointly optimizing the latent codes $z_t$ and network weights $\gamma$ while solving the compressive sensing problem. 
In other words, we can overcome the range limitation of the generator by optimizing generator parameters alongside latent code to get a good reconstruction from compressive measurements as well as good representative latent codes for the video sequence even though the network is highly overparameterized. The resulting optimization problem can be written as 
\begin{equation} 
    \hat z_1,\ldots,\hat z_T; \hat \gamma=\argmin_{z_1,\ldots, z_T;\gamma}\;\sum_{t=1}^T \|y_t-A_t G_\gamma(z_t)\|^2_2,\label{eq_joint}
\end{equation}
where the reconstructed video sequence can be generated using the estimated latent codes and generator weights as $\hat x_t = G_{\hat \gamma}(\hat z_t)$. 

This joint optimization of latent code and generator parameter offer the optimization problem a lot of flexibility to generate a wide range of images. As the generator function is highly non-convex, we initialize $\gamma$ with the pretrained set of weights. After every gradient descent update of the latent codes, $z_t$, we update the model parameters with stochastic gradient descent.

\subsection{Similarity and Low Rank Constraints}

\subsubsection{Similarity Constraints}
A generative prior gives us an opportunity to utilize the corresponding latent codes. The latent codes can be viewed as nonlinear, low-dimensional projection of the original images. In a video sequence, each frame has some similarities with the neighboring frames. Even though the similarity may seem very complex in original dimension, it can become much simpler when we encode each image to a low dimensional latent code. If the latent code is long enough to encode the changes in the image domain, then they can also be used for applying similarity constraint on the image domain.

We assume that if the images are \textit{similar} to each other, then their corresponding latent codes must be similar too. To exploit this structure, we propose to reconstruct the following optimization problem with similarity constraints: 
\begin{equation}
    \min_{z_1,\ldots,z_T;\gamma}\; \lambda \sum_{t=1}^{T}\|y_t-A_t G_\gamma(z_t)\|_2^2+ (1-\lambda )\sum_{t=1}^{T-1}\beta_t\|z_{t+1}-z_t\|_2^2
    \label{eqn_sim}
\end{equation}
where $0<\lambda<1$ and the $\beta_t$ are the weights that represent some measure of similarity between $t^{th}$ and $(t+1)^{th}$ frames. Assuming the adjacent frames in a sequence are close to each other, we fix $\beta_t=1$ for all $t$ for simplicity.

\subsubsection{Low Rank Constraint}
To further exploit the redundancies in a video sequence, we assume that the variation in the sequence of images are localized and the latent codes sequence can be represented in a much lower dimensional space compared to their ambient dimension. For each minibatch, we define a matrix $Z$ such that
\[
   Z=
   [z_1~z_2~ \ldots  ~z_T]
\]
where $z_t$ is the latent code corresponding to $t^{th}$ image of the sequence. To explore low rank embedding, we solve the following constrained optimization:
\begin{equation}\label{eq:lowrankOpt}
    \begin{aligned}
     &\min_{z_1,\ldots,z_T;\gamma}\;\sum_{t=1}^T\|y_t-A_t G_\gamma(z_t)\|_2^2~~\\
     & \;\;\;\;\text{s.t.}\;\;\; \text{rank}(Z)=r. 
    \end{aligned}
\end{equation}

We implement this constraint by reconstructing $Z$ matrix from its top $r$ singular vectors in each iteration. Thus the rank of $Z$ matrix formed by a sequence of images becomes $r$, which implies that we can express each of the latent codes in terms of $r$ orthogonal basis vectors. For rank$(Z)=r$ embedding, we represent each latent code $z$ as a linear combination of the $r$ orthogonal basis vectors $u_1,\ldots, u_r$ as
\begin{equation}
z_i=\sum_{j=1}^{r} \alpha_{ij}u_{j}
\label{eq_interp}
\end{equation}
where $\alpha_{ij}$ is the weight of the corresponding basis vector.

We can now represent a video sequence with $T$ frames with $r$ orthogonal codes. This offers an additional compression to our latent codes.\Rdelete{ We use the same idea to demonstrate that linear interpolation between two latent codes can also generate the intermediate frames in a video sequence.} \Redit{ We use the same idea to linearize motion manifold in latent space.}

\renewcommand{\algorithmicrequire}{\textbf{Input:}}
\renewcommand{\algorithmicensure}{\textbf{Output:}}

\begin{algorithm}
   \caption{Generative Models for Low Rank Representation and Recovery of Videos}
   \label{alg:low_rank}
    \begin{algorithmic} 
        \Require Measurements $y_t$, measurement matrices $A_t$, pretrained generator $G_\gamma(\cdot)$ 
        \State Initialize the latent codes $z_t$.
        \Repeat
        \State Compute gradients w.r.t. $z_t$ via backpropagation.
        \State Update latent code matrix $Z = [z_1~\cdots~z_T]$. 
        \State Threshold $Z$ to a rank-$r$ matrix via SVD or PCA.
        \Statex
        \State Compute gradients w.r.t. $\gamma$ via backpropagation.
        \State Update network weights $\gamma$. 
        \Until{convergence or maximum epochs}
        \Ensure Latent codes: $z_1,\ldots, z_T$ and network weights:  $\gamma$
    \end{algorithmic}
    
\end{algorithm}

\section{Experiments}
In this section, we describe our experimental setup.

\textbf{Choice of generator:} We follow the well-known DCGAN framework \cite{radford2015unsupervised} for our generators except that we
do not use any batch-normalization layer because gradient through the batch-normalization layer is dependent on the batch size and the distribution of the batch. As shown in Figure~\ref{fig:gen}, in DCGAN generator framework, we project the latent code, $z$, to a larger vector using a fully connected network and then reshape it so that it can work as an input for the following deconvolutional layers. Instead of using any pooling layers, in DCGAN framework, authors \cite{radford2015unsupervised} propose strided convolution. All the intermediate deconvolution layers are followed by ReLU activation. The last deconvolution layer is followed by Tanh activation function to generate the reconstructed image $x=G(z)$. 

\textbf{Initial generator training:} 
We train our generators by jointly optimizing the generator parameters, $\gamma$ and latent code, $z$ using SGD optimization by following the procedure in \cite{Bojanowski2018OptimizingTL}. In each iteration, we first update the generator parameters and then update the latent code using SGD. We use squared-loss function, $\ell_2 (x,\hat x)=\|x-\hat x\|_2^2 $ to train the generators. We keep the minibatch size fixed at 256. We use two different trained generators for our experiments: one for RGB images and another for grayscale images. The RGB image generator is trained on CIFAR10 training dataset resized to $64\times64$. We choose CIFAR10 because it has 10 different categories of images, which helps increase the range of the generator. The grayscale image generator is trained on MNIST digit training dataset resized to $64\times64$. We used SGD optimizer for optimizing both latent code and network weights. The learning rate for updating $z$ is chosen as 1 and learning rate for updating $\gamma$ as $0.01$.

\textbf{Measurement matrix:} We used three different measurement matrices in our experiments. We first experiment with original images (i.e., $A_t$ is an identity matrix) to test which sequences can be generated by latent code optimization and which ones require joint optimization of latent codes and network weights. Then we experiment with compressive measurements, for which we choose the entries of the $A_t$ independently from  $\mathcal{N}(0,\frac{1}{m})$ distribution. For a video sequence of $T$ frames, we generate $T$ independent measurement matrices. Then we experiment with missing pixels (also known as image/video inpainting problem) to show that our algorithm works on other inverse problems as well. For experiments with missing pixels, we randomly dropped a fraction of the pixels from each frame.

\textbf{Datasets:} We test our hypothesis on five datasets, which includes both synthetic and real video sequences. The first test set consists of 10 MNIST test digits. We rotate each digit by $2^{\circ}$ per frame for a total of 32 frames. Second test set includes 10 Moving MNIST test sequences \cite{srivastava2015unsupervised}. Each test sequence has 20 frames. For the third test set, we generate a color wheel with 12 colors by dividing a circle into 12 equal slices.  We rotate the color wheel by $1^{\circ}$ per frame for 64 frames. Finally we experiment on different real video sequences from publicly available KTH human action video dataset \cite{schuldt2004recognizing} and UCF101 dataset \cite{soomro2012ucf101}. We show the results on a person walking video from KTH dataset in this paper because of its simplicity. We cropped the video in the temporal dimension to select 80 frames, which show only unidirectional movement. We also show results for an archery video sequence from UCF101 dataset.

\textbf{Performance metric:} We measure the performance of our recovery algorithms in terms of the reconstruction error PSNR. For a given image $x$ and its reconstruction $\hat x$, PSNR is defined as
\begin{equation*}
    \text{PSNR}(x,\hat x)= 20 \log_{10} \frac{\max(x)-\min(x)}{\sqrt{\text{MSE}(x,\hat x)}}
\end{equation*}
where max and min corresponds to the maximal and minimal value the image $x$ can attain respectively, and MSE is the mean squared error.


\section{Results}

\subsection{Compact Video Representation} 
In our first set of experiments, we simply generate a given video sequence using our network by optimizing only over the latent codes and by optimizing jointly over the latent codes and network parameters. In other words, $A_t$ is an identity matrix in these experiments. A summary of our experimental results is presented in Table~\ref{table:org} that correspond to the case when original video sequence is used to estimate latent codes that provide best approximation of the sequence. We observe from Table \ref{table:org} that adding similarity and low-rank constraints provides small improvement in the image approximation performance. This might be because of the fact that the frames are already slowly changing \Redit{and we have enough measurements to approximate them}. However, jointly optimizing both latent codes and network parameters provides a significant gain in the reconstruction PSNR. 

\begin{table*}[t]
	\centering
	\caption{Results for compact video representation via generative model in terms of PSNR. In each experiment, we approximated a video sequence by either optimizing over latent codes or joint optimization over latent codes and network weights. \Redit{First column (Update $z_t$) corresponds to the algorithm of \cite{bora2017compressed}}}
	\label{table:org}
	\vskip 0.15in
	\begin{tabular}{lccc|ccc}
		\hline
		& \multicolumn{3}{c|}{\begin{tabular}[c]{@{}c@{}}Latent code optimization\end{tabular}}                                                                              & \multicolumn{3}{c}{\begin{tabular}[c]{@{}c@{}}Joint  optimization\end{tabular}}                                                                                         \\ \hline
		& \multicolumn{1}{c}{\begin{tabular}[c]{@{}c@{}}Update $z_t$\end{tabular}} & \multicolumn{1}{c}{\begin{tabular}[c]{@{}c@{}}Low-rank \\ constraints ($r=5$) \end{tabular}} & \multicolumn{1}{c|}{\begin{tabular}[c]{@{}c@{}}Similarity \\ constraint\end{tabular}} & \multicolumn{1}{c}{\begin{tabular}[c]{@{}c@{}}Update \\ $z_t$ and $\gamma$\end{tabular}} & \multicolumn{1}{c}{\begin{tabular}[c]{@{}c@{}}Low-rank \\ constraints ($r=5$)\end{tabular}} & \multicolumn{1}{c}{\begin{tabular}[c]{@{}c@{}} Similarity\\ constraint\end{tabular}} \\ \hline
		Rotating MNIST & 25.82 &   25.73   & 26.81  & 33.75   & 33.78      & 33.9     \\
		Moving MNIST   & 18.55   & 16.99 & 18.51& 31.17 & 31.16 & 31.15   \\
		Color Wheel    & 18.24 & 17.97  & 18.31 & 22.07 & 21.92 & 22.05  \\
		Archery        & 24.15& 23.13       & 24.49 & 26.5 & 23.15 & 27.26\\ 
		Person Walking  & 27.55 &  23.30 &27.55  &27.9  & 26.72 &27.91 \\
		\bottomrule
	\end{tabular}
\end{table*}

\subsection{Optimization over $z_t$ with Constraints} 
\begin{figure}[!h]
\centering
\includegraphics[width=0.7\linewidth]{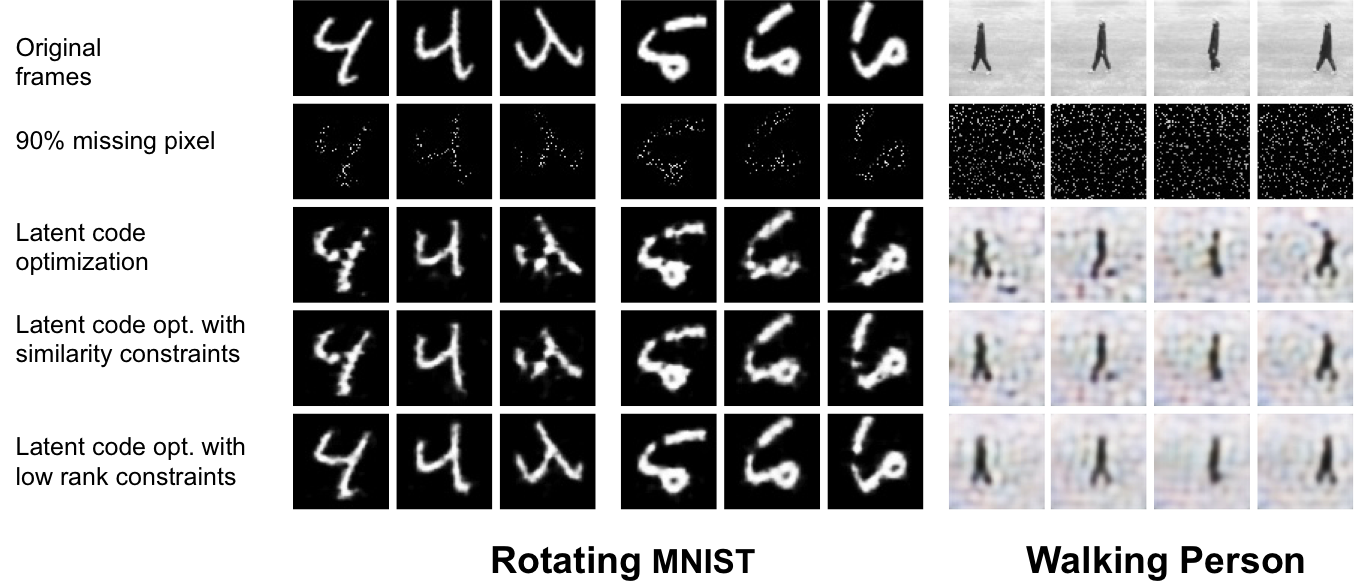}

\caption{Example reconstruction results from inpainted video sequence with 90\% missing pixels for two rotating MNIST sequences and person walking video sequence using latent code optimization.}
\label{rec_miss_latent}
\end{figure}

In our first experiment, we test latent code optimization with and without similarity and low-rank constraints. We show some example reconstructions for the inpainting problem with 90\% missing pixels in Figure~\ref{rec_miss_latent}. For similarity constraint, $\lambda = 0.6$ is chosen for both cases. For low-rank constraints, the optimal values of rank for Rotating MNIST and Person Walking are $rank=4$ and $rank=16$, respectively. We can observe for very low measurements, low rank generator not only represent the video sequence with lower number of parameters in latent codes ($12.5\%$ and $20\%$ of the total frames respectively for Rotating MNIST and Person Walking), it also gives boost in reconstruction performance.

We also performed a number of experiments for latent code optimization (with and without similarity and low-rank constraints) for different datasets and measurements. A summary of our experimental results is presented in Table~\ref{table:inv}. The results refer to experiments in which we estimate latent codes from the compressive measurements of the sequence. We observe that adding similarity or low-rank constraints in the compressive sensing problems  shows significant improvement in the quality of reconstruction.

\begin{table*}[t]
\centering
\caption{Reconstruction PSNR for compressive sensing problems. First four rows correspond to video recovery from $m$ Gaussian measurements. Last five rows correspond to the recovery of videos from 80\% missing pixels per frame. \Redit{First column (Update $z_t$) corresponds to the algorithm of \cite{bora2017compressed} }}
\label{table:inv}
\vskip 0.15in
\begin{tabular}{llll|lll}
\hline
               & \multicolumn{3}{c|}{\begin{tabular}[c]{@{}c@{}}Latent code  optimization\end{tabular}}                                                                              & \multicolumn{3}{c}{\begin{tabular}[c]{@{}c@{}}Joint  optimization\end{tabular}}                                                                                         \\ \hline
               & \multicolumn{1}{c}{\begin{tabular}[c]{@{}c@{}}Update $z_t$\end{tabular}} & \multicolumn{1}{c}{\begin{tabular}[c]{@{}c@{}}Low-rank\\ constraint\end{tabular}} & \multicolumn{1}{c|}{\begin{tabular}[c]{@{}c@{}}Similarity\\ constraint\end{tabular}} & \multicolumn{1}{c}{\begin{tabular}[c]{@{}c@{}}Update \\$z_t$ and $\gamma$\end{tabular}} & \multicolumn{1}{c}{\begin{tabular}[c]{@{}c@{}}Low-rank\\ constraint\end{tabular}} & \multicolumn{1}{c}{\begin{tabular}[c]{@{}c@{}}Similarity\\ constraint\end{tabular}} \\ \hline
& \multicolumn{6}{c} {\it Experiments with compressive Gaussian measurements}\\
Rotating MNIST ($m=200$) & 20.35  & 20.75 (r=5)    & 22.13  &  30.9  &  31 (r=5)   & 32.97     \\
Moving MNIST ($m=512$)   & 16.75   &16.9 (r=12)& 17.57  & 24.43  & 27.03 (r=4)  &  27.2   \\
Color Wheel ($m=1024$)    & 16.95 & 17.96 (r=6)  &17.09  & 21.92& 23.71 (r=6) & 21.8  \\
Archery ($m=512$)       &21.58 &  23.54(r=16)      & 23.15 & 25.82& 26.9 (r=21) &25.83\\ 
\midrule
&\multicolumn{6}{c} {\it Experiments with 80\% Missing pixels}\\
Rotating MNIST & 19.15  & 25.07(r=4) &24.45  & 26.54 &29.58 (r=3) &28.53\\
Moving MNIST  & 16.44 &16.82 (r=9)  & 17.34 &18.65 &19.02(r=9) &19.55\\
Color Wheel   & 16.54 & 17.85 (r=6) & 16.75 &18.46 &19.96 (r=4) &18.88\\
Archery   & 23.15 &23.8 (r=22)  & 23.32 &23.6 & 23.81 (r=21) &23.57\\
Person Walking   & 25.34  & 26.1 (r=21)  & 25.9 &25.8  & 26.17 (r=22) & 25.96 \\
\bottomrule

\end{tabular}
\end{table*}

\subsection{Joint Optimization over $z_t$ and $\gamma$ with Constraints}

As we discussed before in Figure~\ref{mov_mnist}, the joint optimization over $z_t$ and $\gamma$ can generate images that are very different from the images network is trained on. Table~\ref{table:org} refers to similar experiments in which we are given the original video sequence and we want to estimate latent codes and network weight that can best approximate the given video sequence. We observe that joint optimization offers a significant performance boost compared to latent code optimization alone. As we discussed before, this is expected because we have a lot more degrees of freedom in the case of joint optimization than what we have for latent code optimization. The similarity or low-rank constraints do not provide a significant boost while approximating the video sequence. \Rdelete{Adding the similarity constraints in the joint optimization does not show a significant improvement.} 

Table~\ref{table:inv} summarizes results for compressive measurements, where we are only given linear measurement of the video sequence and we want to estimate the latent codes $z_t$ and network weights $\gamma$ that minimize the objectives in \eqref{eq_joint} or \eqref{eq:lowrankOpt}. 
We performed experiment on image inpainting and compressive sensing problems. For image inpainting problem, we show reconstruction results for $80\%$ missing pixels in Table \ref{table:inv}. We also show results for different compressive measurements for different synthetic and real video sequences. We can observe from Table \ref{table:inv} that with low-rank constraints on the generator,  we can not only represent the whole video sequence with a very few latent codes, \Rdelete{we} \Redit{but also get} better reconstruction than full rank cases. Similarity constraint on latent codes also show improvement in reconstruction performance when the measurements are low.

\begin{figure*}[!h]
\centering
    
   \includegraphics[width=1\textwidth]{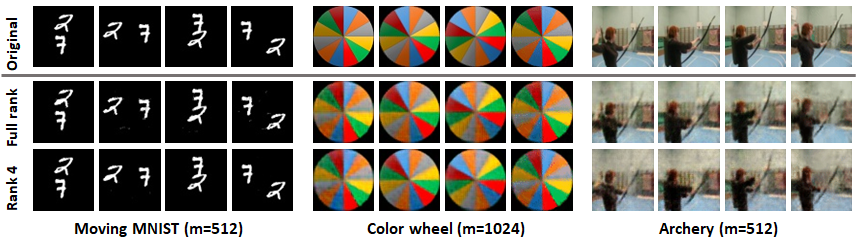}
   \caption{Examples of reconstructed images for experiments with different datasets using linear measurements. The results are from joint optimization of latent code and generator weights. First row shows samples from original video sequence. Second row shows reconstruction without low-rank constraint. Third row shows results when latent codes for each sequence are restricted to a rank-4 matrix. }
   \label{fig:jointImages}
\end{figure*}

Some examples of video sequences from compressive measurements are presented in Figure~\ref{fig:jointImages}. In each of the experiments, we compute $m$ Gaussian measurements of each frame in a sequence and then solve the optimization problems in \eqref{eq_joint} (this corresponds to the full-rank recovery) and \eqref{eq:lowrankOpt} with $r=4$ (this corresponds to the rank-4 recovery). We observe that low-rank constraints provide a small improvement in terms of the quality of reconstruction.


\subsection{Linearizing Motion Manifold via Joint Optimization \Redit{and Low Rank Constraint} }
In this section, we present our preliminary experiments on linearizing articulation manifold of a video sequence by imposing low-rank structure on the latent codes. In our experiment, we force our latent codes to map on a straight line by defining each $z_t = \bar z + \alpha_t u$, where $\bar z, u \in \mathbb{R}^k$ and $\alpha_t \in \mathbb{R}$ are scalar. We impose this rank-2 constraint by solving the problem in \eqref{eq:lowrankOpt} but instead of approximating the $z_t$ using the top two singular vectors, we approximate them using their mean and first principal vector. 

\Redit{We further investigate the linearization of multiple video sequences while optimizing the same generator weights to generate those sequences. In this experiment, we form the $Z$ matrix by concatenating latent codes for multiple different sequences. Then we apply rank-2 constraint on the entire $Z$ matrix using top two singular vectors. We simultaneously apply linearity constraint on each sequence by imposing rank=2 constraint on the latent codes for each sequence separately using mean and first principal vector as mentioned above.}

\begin{figure*}[!h]
\centering
   \begin{subfigure}[t]{0.31\linewidth}
   \includegraphics[width=1\textwidth]{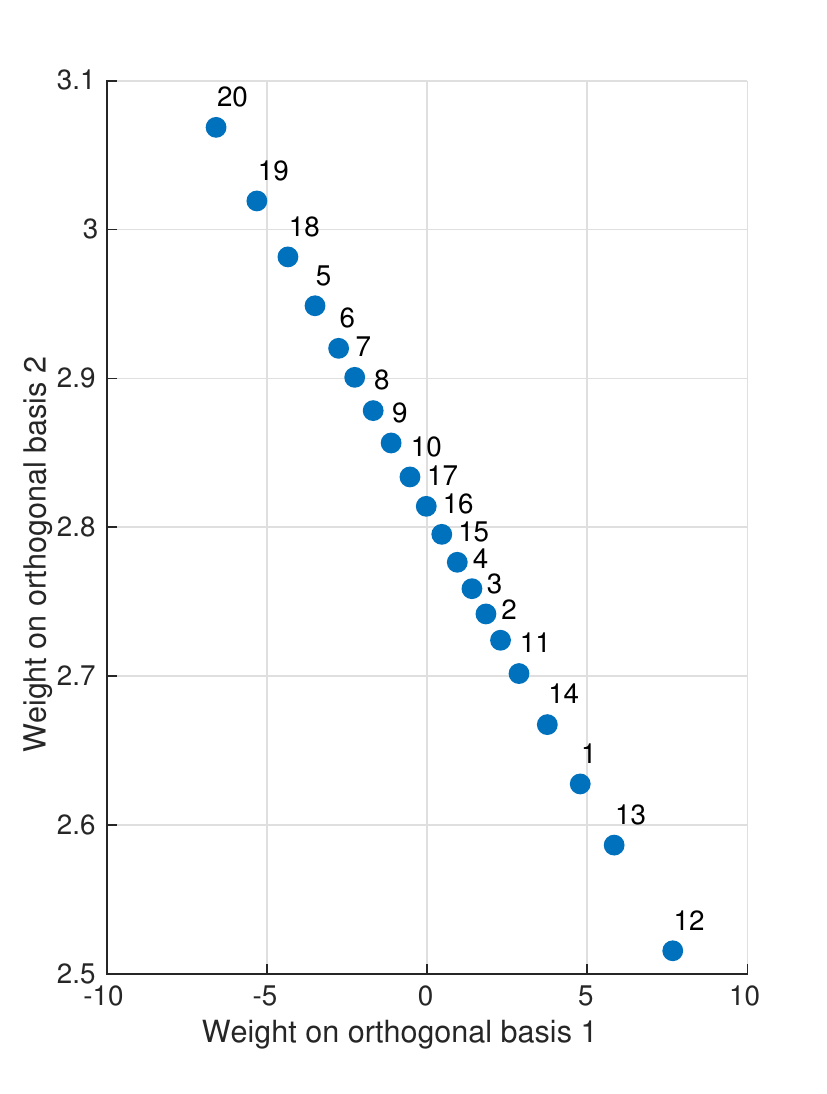}
   \caption{Manifold of latent codes for Moving MNIST.}
   \label{z_manifold_mov}
   \end{subfigure}\hfill
      \begin{subfigure}[t]{0.31\linewidth}
   \includegraphics[width=1\textwidth]{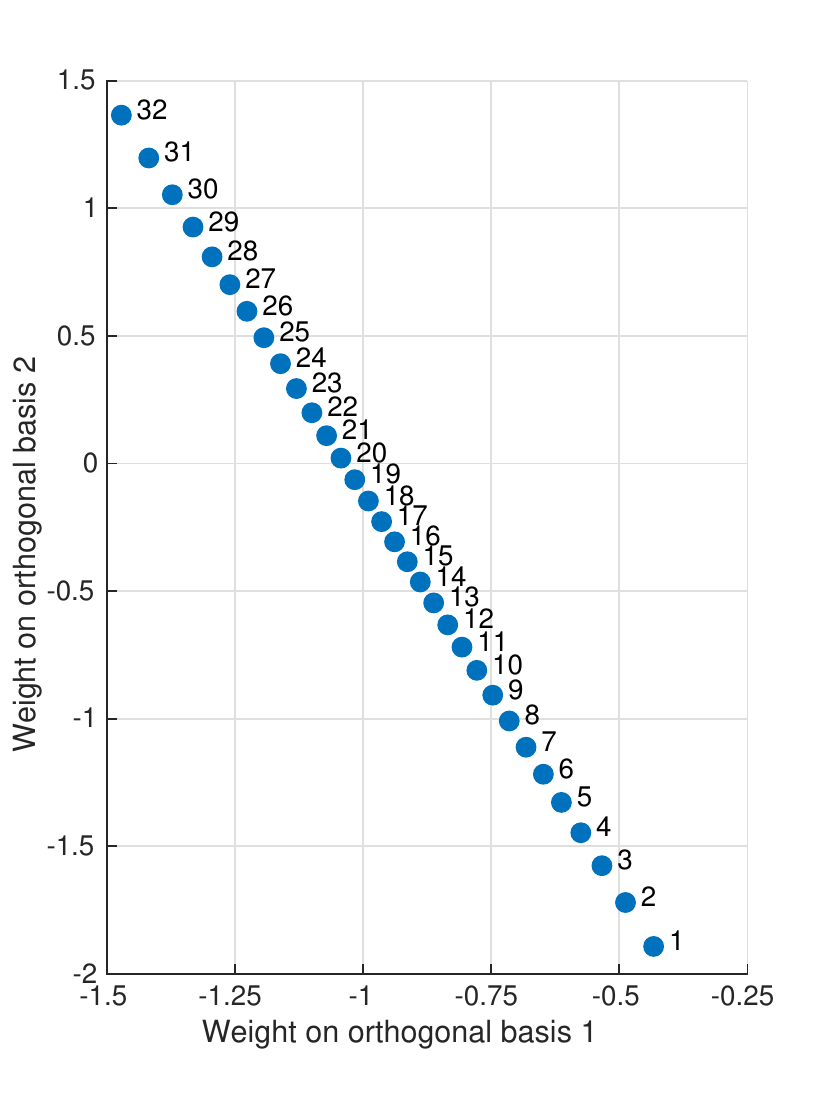}
   \caption{Manifold of latent codes for Rotating MNIST.}
   \label{z_manifold_rot}
   \end{subfigure}\hfill
   \begin{subfigure}[t]{0.32\linewidth}
   \includegraphics[width=1\textwidth]{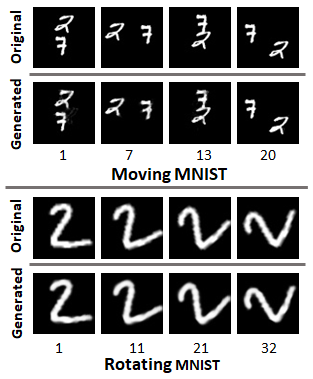}
   \caption{Generated images for some frames of the sequences. We denote the frame number below each frame.}
   \label{z_int}
   \end{subfigure}
   \caption{\Rdelete{Generated} \Redit{Approximated} images for Moving MNIST and one Rotating MNIST video sequences  \Redit{ from original video sequences using joint optimization} in which the latent codes are constrained to lie on a straight line using PCA. In (a) and (b), these representations are linear by virtue of the constraint. Furthermore, for Rotating MNIST, the latent codes are sequentially arranged. The quality of generated frames are good (Average PSNR for Moving MNIST: 26.8 dB; Rotating MNIST: 33.6 dB).}\label{fig:2dEmbedding}
\end{figure*}

\begin{figure*}[!h]
\centering
   \begin{subfigure}[t]{0.6\linewidth}
   \includegraphics[width=1\textwidth]{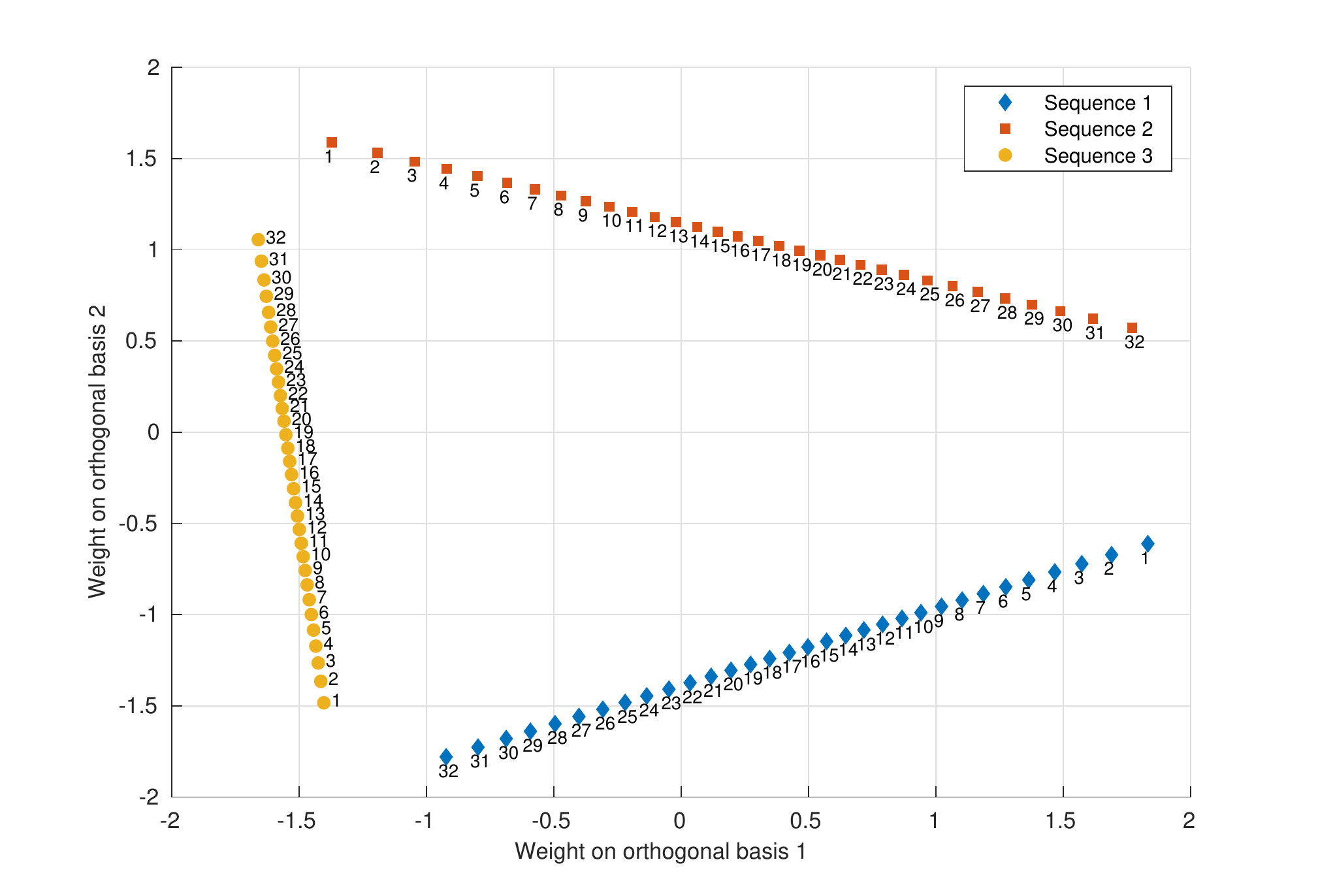}
   \caption{Manifold of latent codes for 3 sequences from Rotating MNIST.}
   \label{fig:z_manifold_rot_3seq}
   \end{subfigure}
   \begin{subfigure}[t]{0.21\linewidth}
   \includegraphics[width=1\textwidth]{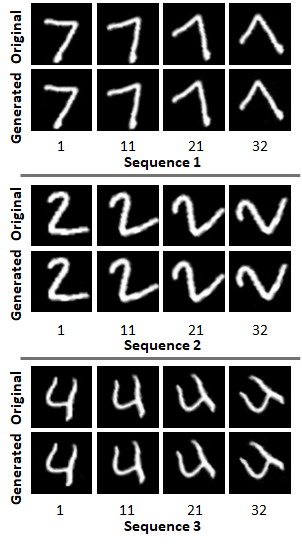}
   \caption{Generated images for frames of 3 sequences from Rotating MNIST.}
   \label{fig:rec_rot_mnist_3seq}
   \end{subfigure}
  \caption{\Redit{Approximated images and latent space representation from original video sequences for three different Rotating MNIST video sequences using joint optimization of latent codes and network weights for the same generator. Here the latent codes of each sequences are constrained to lie on a straight line using PCA. Different sequences are aligned to different lines in 2D plane. Furthermore, they maintained sequential arrangement.}} \label{fig:2dEmbedding_3seq}
\end{figure*}

We plot the embedding of each $z$ in terms of two orthogonal basis vectors in Figures~\ref{z_manifold_mov} and \ref{z_manifold_rot}. We observe that a well-defined rotation in image domain is translated into a line in the latent space. We also observe that as we increase the rotating angles, the corresponding embedding moves along a straight line in the direction of first principal vector in an increasing order. \Redit{We plot the embedding of three sequences of Rotating MNIST in Figure \ref{fig:z_manifold_rot_3seq}. We observe that the rotation of different digits are translated into different lines in the 2D latent space. Furthermore, latent codes for each of the sequences preserves their sequential order.} However, in the case of moving MNIST, even though we get perfect reconstruction with the line embedding, but the order of the video sequence is not preserved in the embedded space. We did not impose any constraint in our optimization to preserve the order, but we expect that if the video sequence changes in such a manner that frames that are farther in time are also farther in content, then we will see the order will be preserved. We leave this investigation for future work.


\subsection{Interpolation in Latent Space to Generate  Missing Frames}
\begin{figure*}[!h]
	\centering
	\begin{subfigure}[t]{0.6\linewidth}
		\includegraphics[width=1\textwidth]{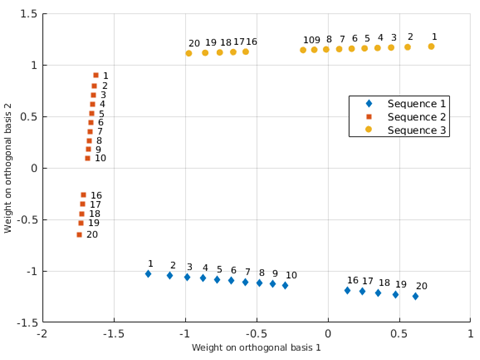}
		\caption{Manifold of latent codes for 3 sequences from Rotating MNIST with missing frames.}
		
		\label{fig:z_manifold_interp_3seq}
	\end{subfigure}\hfill
	\begin{subfigure}[t]{0.32\linewidth}
		\includegraphics[width=1\textwidth]{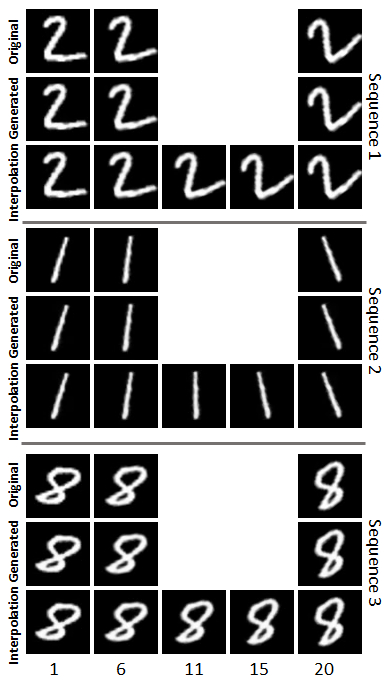}
		\caption{Approximation and interpolation result for some frames.}
		
		\label{fig:rec_interp_3seq}
	\end{subfigure}
	\caption{Approximation and interpolation (in latent space) images and corresponding latent space representation for 3 Rotating MNIST sequences with missing frames. We denote the frame number for each frame. Frame 11 to Frame 15 are missing while joint optimization of $z$ and $\gamma$. } 
	\label{fig:interp_3seq}
\end{figure*}

If the latent codes follow some sequential order, it is possible to generate intermediate images between each frames. We test this idea using three Rotating MNIST sequences. Each sequence originally contained 20 frames, where, in each frame, the digit is rotated $2^\circ$ from the previous frame. However, we set aside $11^{th}$ to $15^{th}$ frames while optimizing the generator to approximate those frames. We perform joint optimization of $z$ and $\gamma$ using rank=2 constraint on the latent codes and linearization constraint on the latent codes of each sequence. When we observe the latent code representation for the approximated images, we observe that the latent codes follow sequential order but there are significant gap between the latent codes of $10^{th}$ and $16^{th}$ frames. We can observe this phenomena in Figure \ref{fig:z_manifold_interp_3seq}. We then try to generate 1000 frames between frame 1 and frame 20 using linear interpolation between corresponding latent codes. We keep the same network weights which is giving us the approximation of the original sequences. We can observe from Figure \ref{fig:rec_interp_3seq} that we can generate the missing frames in that way. However, we can choose frame 1 and 20 here as two end points for linear interpolating because the entire sequence is maintaining the sequential order in their linear latent space representation. But in cases where the sequence only maintains sequential order locally, we can select interpolation end points from the cluster of frames which maintains sequential order.

\subsection{Low-Dimensional Embedding of Complex Motion}
\begin{figure}[h]
	\centering
	\begin{subfigure}[t]{0.50\linewidth}
		\includegraphics[width=1\textwidth]{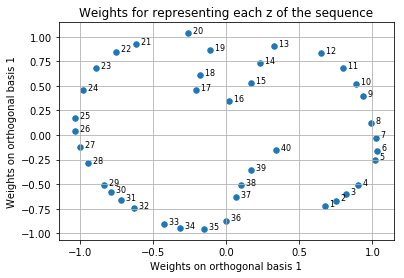}
		\caption{Latent code representation for approximated rotating bunny sequence.}
		\label{fig:bunny_manifold}
	\end{subfigure}\hfill
	\begin{subfigure}[t]{0.4\linewidth}
		\includegraphics[width=1\textwidth]{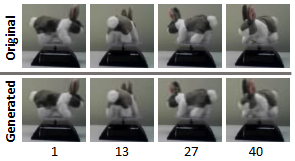}
		\caption{Approximated images for rotating bunny sequence.}
		\label{fig:bunny_rec}
	\end{subfigure}

	\caption{Approximated images and corresponding latent space for rotating bunny sequence. We constrain the latent codes to lie on 2D plane of a 3D space using mean and first two principal vectors. }
	\label{fig:bunny}
\end{figure}

We further experiment on a complex real life motion using spinning figures dataset from \cite{lederman2018learning}. We selected a rotating bunny sequence and cropped only the bunny from the images. The bunny completes one rotation in 15 frames. We selected first 10 frames from each of the 4 full rotations and keep the similar rotations close to each other. We try to find out if this sequence maintain its sequential order in any latent space. We observe the representation of the sequence in latent space using $rank=3$ constraint. We impose $rank=3$ constraint by selecting mean and first two principal vectors. So, the latent codes are constrained to 2D plane in the 3D space. We show the approximation of bunny sequence using this constraint in Figure \ref{fig:bunny_rec} and the corresponding latent space representation in Figure \ref{fig:bunny_manifold}. We can observe from the latent space representation that the sequence maintained its sequential order in this representation.

\section{Discussion and Future Work}

We proposed a generative model for low-rank representation and reconstruction of video sequences. We presented experiments to demonstrate that video sequences can be reconstructed from compressive measurements by either optimizing over the latent code or jointly optimizing over the latent codes and network weights. We observed that adding similarity and low-rank constraints in the optimization regularizes the recovery problems and improves the quality of reconstruction. We presented some preliminary experiments to show that low-rank embedding of latent codes with joint optimization can potentially be useful in linearizing articulation manifolds of the video sequence. \Redit{An implementation of our algorithm with pretrained models is available here: \href{https://github.com/CSIPlab/gmlr}{https://github.com/CSIPlab/gmlr}.}

In all our experiments, we observed that joint optimization performs remarkably well for compressive measurements as well. Even though the number of measurements are extremely small compared to the number of parameters in $\gamma$, the solution almost always converges to a good sequence. We attribute this success to a good initialization of the network weights and hypothesize that a ``good set of weights'' are available near the initial set of weights in all these experiments. We intend to investigate a proof of the presence of good local minima around initialization in our future work.

\bibliography{ref}
\bibliographystyle{abbrv}

\pagebreak
\appendix

\section{Supplementary Material}
\subsection{Image Inpainting on Additional Video Sequences}
\begin{figure*}[!h]
\centering
   \begin{subfigure}[t]{1\linewidth}
   \includegraphics[width=1\textwidth]{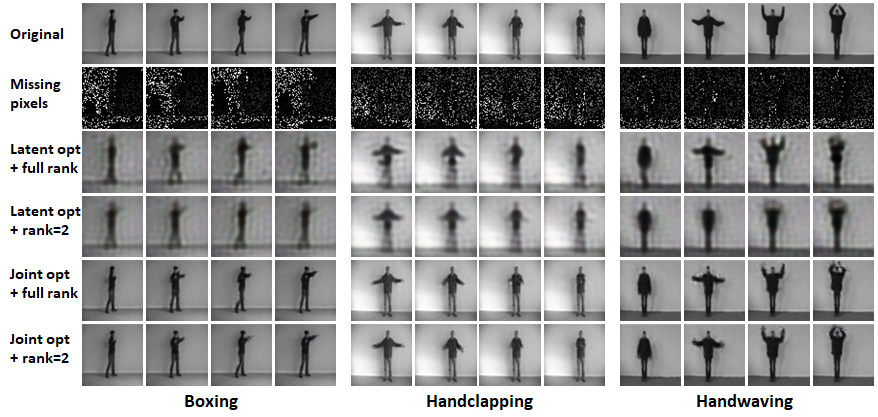}
   \end{subfigure}
   
      \begin{subfigure}[t]{1\linewidth}
   \includegraphics[width=1\textwidth]{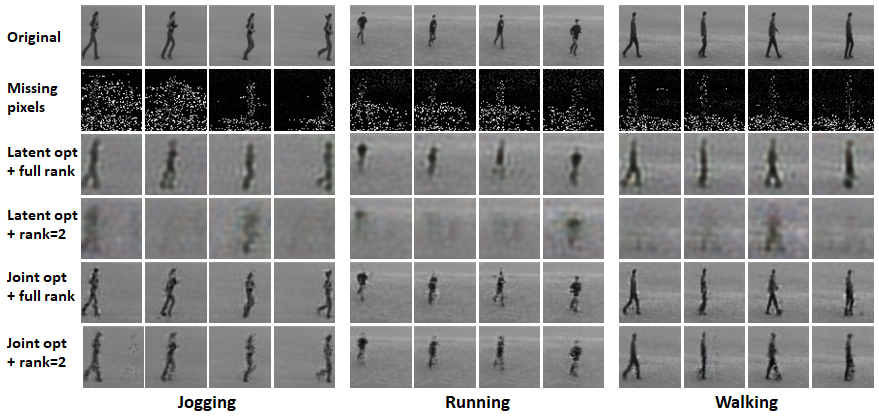}
   \end{subfigure}
     \caption{Reconstructions from inpainted video sequence with 80\% missing pixels for different videos from KTH dataset. We enforce rank=2 with linearization constraint using PCA. }
   \label{fig:kth_miss}
\end{figure*}

We experiment on different video sequences from all six categories (Boxing, Handclapping, Handwaving, Jogging, Running, Walking) from KTH video dataset. To reduce computational complexity, we have selected part of these videos in a batch. Table \ref{table:kth_miss} includes the number of frames for our test videos. We experiment on image inpainting with 80\% missing pixels. We experiment for both latent code optimization and joint optimization of latent code and network weight. In Table \ref{table:kth_miss}, we report experimental results and in Figure \ref{fig:kth_miss}, we demonstrate some representational examples. Joint optimization of $z$ and $\gamma$ significantly outperforms latent code optimization because the video sequences are not from the range of the pretrained generator. Furthermore, applying rank=2 linearization constraint on latent code we observe similar performance as full rank reconstruction for joint optimization.

\begin{table}[!h]
\centering
\caption{Reconstruction PSNR for inpainting problem with 80\% missing pixels on different KTH video sequences. We show results for latent code optimization and joint optimization of $z$ and $\gamma$ with and without linearization constraint on latent codes.}
\label{table:kth_miss}
\vskip 0.15in
\begin{tabular}{llll|ll}
\hline
\multirow{2}{*}{Video} & \multicolumn{1}{c}{\multirow{2}{*}{\begin{tabular}[c]{@{}c@{}}No. of\\ Frames\end{tabular}}} & \multicolumn{2}{c|}{Latent Code optimization}                            & \multicolumn{2}{c}{Joint Optimization}                                 \\ \cline{3-6} 
                       & \multicolumn{1}{c}{}                                                                         & \multicolumn{1}{c}{Full rank} & \multicolumn{1}{c|}{Rank=2 (linearized)} & \multicolumn{1}{c}{Full rank} & \multicolumn{1}{c}{Rank=2 (linearized)} \\ \hline
Boxing                 & 50                                                                                           & 22.45                         & 22.62                                    & 32.37                         & 30.38                                   \\ \hline
Handclapping           & 50                                                                                           & 26.03                         & 26.2                                     & 35.74                         & 33.98                                   \\ \hline
Handwaving             & 50                                                                                           & 22.29                         & 20.65                                    & 30.01                         & 27.48                                   \\ \hline
Jogging                & 30                                                                                           & 23.82                         & 18.58                                    & 26.4                          & 24.01                                   \\ \hline
Running                & 30                                                                                           & 25.74                         & 20.66                                    & 27.54                         & 27.56                                   \\ \hline
Walking                & 55                                                                                           & 23.72                         & 18.44                                    & 27.53                         & 27.33                                   \\ \hline
\bottomrule
\end{tabular}
\end{table}

\newpage

\subsection{Untrained Network vs Pretrained Network for Initialization}

\begin{figure*}[!h]
\centering
    
   \includegraphics[width=1\textwidth]{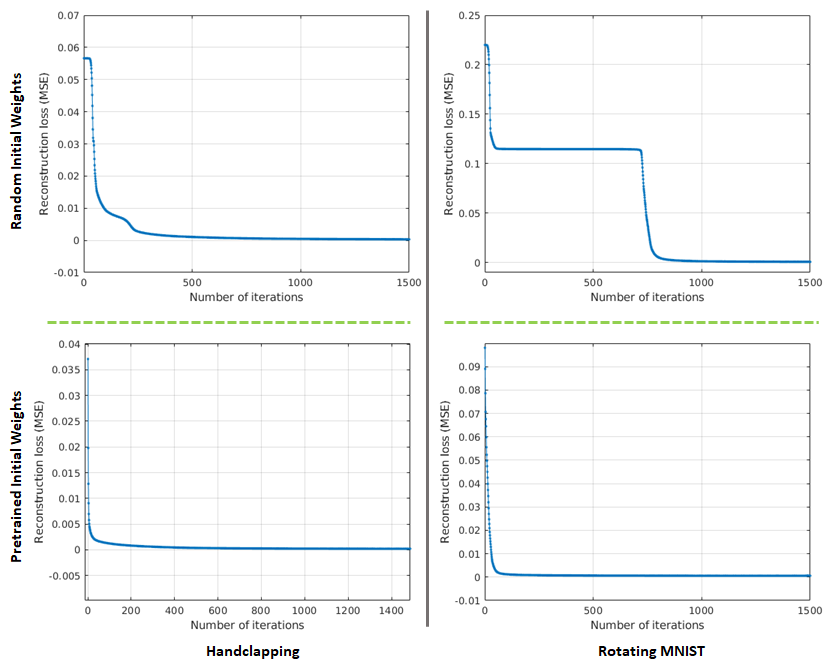}
   \caption{ Reconstruction loss curve for a Handclapping and a Rotating MNIST video for initialization with untrained network weights and pretrained network weights. The loss curves are for inpainting experiments with 80\% missing pixels.}
   \label{fig:loss_init}
\end{figure*}

We further experiment with untrained generator like \cite{Ulyanov2017DeepIP, van2018compressed}. We observe that if we initialize network weights with pretrained network weights, network converges faster even for the images that were not used in the training but fall under the similar distribution. In Figure\ref{fig:loss_init}, we show reconstruction loss vs number of iteration curve for a Rotating MNIST and a Handclapping video. We show these results for inpainting problem with 80\% missing pixels. We can observe that for Rotating MNIST video, random initialization shows false convergence before finally converging. It becomes difficult for some datasets like Moving MNIST to find a convergence using untrained network weights as initialization. So, we use the weights of a pretrained network as initialization.

\subsection{Network Parameters}

We use two different generator networks for RGB image generation and grayscale image generation. In both generators, we use $4\times4$ filters in deconvolutional layers. For RGB image generator, a 256 dimensional latent code is projected and reshaped into $512\times4\times4$ whereas for grayscale image generator, a 32  dimensional latent code is projected and reshaped into $256\times4\times4$. The number of kernel for each deconvolutional layer of RGB image generator is 256, 128, 64 and 3, respectively. For grayscale image generator, number of kernel for each deconvolutional layer is 128,64,32 and 1, respectively. The number of parameters for each generator is shown in Table \ref{table:param}.

\begin{table}[!h]
\centering
\caption{Number of parameters in different layers of the generator networks used in the experiments.}
\label{table:param}
\vskip 0.15in
\begin{tabular}{lll}
\hline
\multicolumn{1}{c}{\multirow{2}{*}{Layers}}          & \multicolumn{2}{c}{Number of Parameters}                                                 \\ \cline{2-3} 
\multicolumn{1}{c}{}                                 & \multicolumn{1}{c|}{RGB Image Generator} & \multicolumn{1}{c}{Grayscale Image Generator} \\ \hline
\multicolumn{1}{c}{Fully-connected + reshape + ReLU} & \multicolumn{1}{l|}{2,097,152}             & 131,072                                        \\ \hline
Deconv 1 + ReLU                                      & \multicolumn{1}{l|}{2,097,152}             & 524,288                                        \\ \hline
Deconv 2 + ReLU                                      & \multicolumn{1}{l|}{524,288}              & 131,072                                        \\ \hline
Deconv 3 + ReLU                                      & \multicolumn{1}{l|}{131,072}              & 32,768                                         \\ \hline
Deconv 4 + Tanh                                      & \multicolumn{1}{l|}{3,072}                & 512                                           \\ \hline
\midrule
                                         \textbf{Total Parameters}            & \multicolumn{1}{l}{4,852,736}             & 819,712                                        
\end{tabular}
\end{table}

\end{document}